\documentclass[10pt,twocolumn,letterpaper]{article}

\usepackage[pagenumbers]{cvpr} 

\usepackage{graphicx}
\usepackage{amsmath}
\usepackage{amssymb}
\usepackage{booktabs}
\usepackage{adjustbox}

\usepackage[pagebackref,breaklinks,colorlinks]{hyperref}

\usepackage[capitalize]{cleveref}
\crefname{section}{Sec.}{Secs.}
\Crefname{section}{Section}{Sections}
\Crefname{table}{Table}{Tables}
\crefname{table}{Tab.}{Tabs.}

\def\cvprPaperID{27} 
\def\confName{CVPR}
\def\confYear{2022}

\begin{document}

\title{Multi-Modal Domain Fusion for Multi-modal Aerial View Object Classification}

\author{Sumanth Udupa\\
Department of Aerospace Engineering,\\
Indian Institute of Science, Bangalore\\
{\tt\small sumanthudupa@iisc.ac.in}
\and
Aniruddh Sikdar\\
Robert Bosch Centre for Cyber-Physical Systems,\\
Indian Institute of Science, Bangalore\\
{\tt\small aniruddhss@iisc.ac.in}
\and
Suresh Sundaram\\
Department of Aerospace Engineering,\\
Indian Institute of Science, Bangalore\\
{\tt\small vssuresh@iisc.ac.in}
}

\maketitle

\begin{abstract}
   Object detection and classification using aerial images is a challenging task as the information regarding targets are not abundant. Synthetic Aperture Radar(SAR) images can be used for Automatic Target Recognition(ATR) systems as it can operate in all-weather conditions and in low light settings.
   But, SAR images contain salt and pepper noise(speckle noise) that cause hindrance for the deep learning models to extract meaningful features. Using just aerial view Electro-optical(EO) images for ATR systems may also not result in high accuracy as these images are of low resolution and also do not provide ample information in extreme weather conditions. 
   Therefore, information from multiple sensors can be used to enhance the performance of Automatic Target Recognition(ATR) systems.
   In this paper, we explore a methodology to use both EO and SAR sensor's information to effectively improve the performance of the ATR systems by handling the shortcomings of each of the sensors.
   A novel Multi-Modal Domain Fusion(MDF) network is proposed to learn the domain invariant features from multi-modal data and use it to accurately classify the aerial view objects. 
   The proposed MDF network achieves top-10 performance in the Track-1 with an accuracy of 25.3\% and top-5 performance in Track-2 with an accuracy of 34.26\% in the test phase on the PBVS MAVOC Challenge dataset\cite{low2022multi}.

\end{abstract}

\section{Introduction}
\label{sec:intro}
Automatic Target Recognition(ATR) systems can be used for remote sensing applications like object tracking,traffic monitoring and large scale surveillance.
These systems can also be used for forest fire monitoring and disaster management\cite{akhloufi2021unmanned}\cite{yuan2015survey}.
Deep learning models have made major developments in computer vision related tasks as they generally outperform conventional techniques due to their robust feature extraction capability.
Electro-optical(EO) data is the dominant input for these models as  huge labeled datasets can be created manually using human operators.
Using EO sensor for extended period of time for earth remote sensing applications is not feasible as it cannot capture important information in all-weather and no light conditions.
Synthetic Aperture Radar(SAR) is used to generate high resolution images for such tasks as it can operate in all weather conditions and even in low-light settings.
Since SAR images are not easy to understand or intuitive as shown in figure 1, they cannot be manually annotated accurately.Since its not possible to generate a huge labelled dataset,these data driven deep learning models cannot be directly used for SAR images\cite{9025419}.
Object detection in satellite images is a challenging task as the scale of view is large and the target of interest tend to be small.Due to this,the feature information regarding the target is less.
The targets have only few tens of pixels of information in the satellite image, which is less information available for the convolutional neural network(CNN) to extract meaningful features\cite{yang2019air}.
Discriminative features regarding the target can be obtained from Electro-optical(EO) and SAR sensors, and this multi-modal data can be used to improve the performance of ATR systems.\\
\begin{figure}
    \centering
    \includegraphics{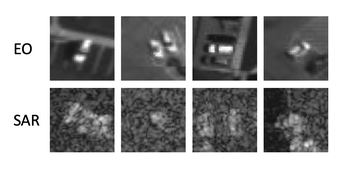}
    \caption{Few EO samples and the corresponding SAR samples from the PBVS MAVOC challenge dataset. }
    \label{fig:dataset}
\end{figure}
Techniques like transfer learning and knowledge distillation have been used to learn the common discriminative features of a target in both the domains.
Yang \textit{et al.}\cite{yang2021cross} proposed a two-way knowledge transfer method between EO and SAR domain by using a teacher-student dual model.
A semi-supervised domain adaptation algorithm was proposed by Rostami \textit{et al.}\cite{9025419} for transferring knowledge from EO to SAR domain.
Data points from both the domains are mapped into a domain invariant space to transfer knowledge across the domains.
Sliced-Wasserstein Distance(SWD) is used to minimize the discrepancy between the source and target distributions and their class conditional densities to make the embedding space domain invariant.\\
Inspired by semi-supervised domain adaptation\cite{9025419} which transfers knowledge from source  to the target domain,we propose a two way knowledge transfer across both  EO and SAR domains.
A twin network is used wherein the images from the two domains are taken as inputs for the network and the outputs are fused together to make the final predictions.
Discriminative features from the labelled and unlabelled multi-modal data  are used to extract domain invariant features in the shared latent space.
A loss function is proposed to train the network in a supervised and semi-supervised learning fashion and the discrepancy between the probability distributions of both the domains are minimized using Sliced-Wasserstein Distance(SWD)\cite{rabin2011wasserstein} in the latent space.
The main contribution of this paper is a Multi-Modal Domain Fusion(MDF) network proposed for multi-modal aerial view object classification using SWD loss function.

Based on the proposed method,we get top-10 performance in both Track-1 and Track-2 in PBVS MAVOC challenge.In Track -1, we get a top-1\% accuracy of 25.3\% on the final test phase data.
In track-2, we get a top-1\% accuracy of 34.26\% on the final test data placing us among the top-5 performing teams in the PBVS MAVOC Challenge.\\
The rest of the paper is organized as follows.Related works in presented in section 2.In section 3,the proposed methodology used for the challenge is explained in detail.Experimental results for both Track-1 and Track-2 are explained in section 4.Finally,the conclusion and future works are discussed in section 5. 
\begin{figure*}[ht]
    \centering
    \includegraphics[width=14cm]{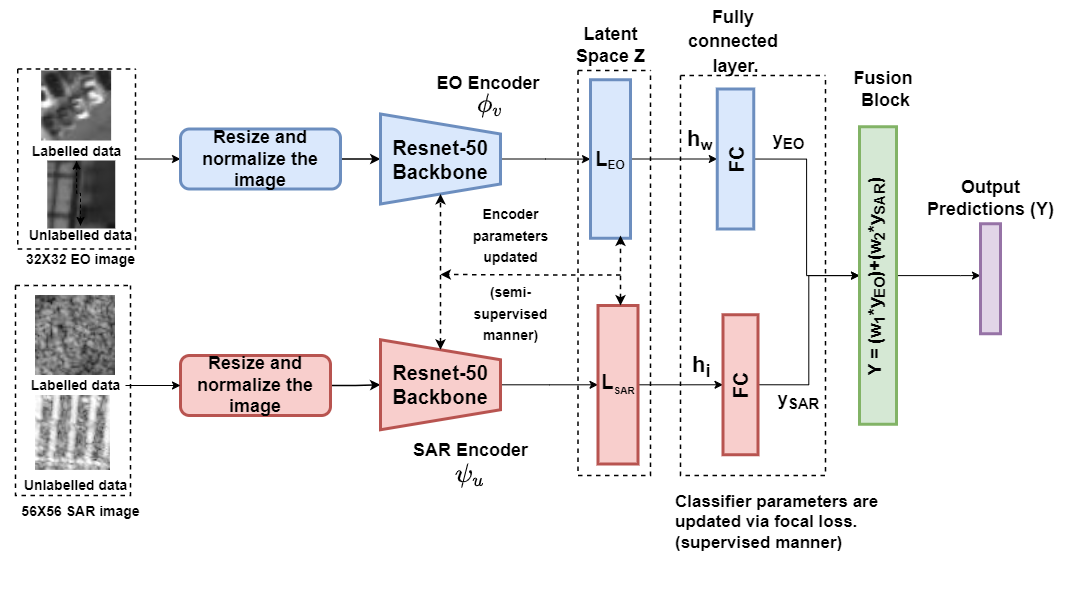}
    \caption{Schematic diagram of the Multi-Modal Domain Fusion Network for Aerial view object Classification. Input to the EO encoder is the labelled and unlabelled EO data. Similarly,input to SAR encoder is the labelled data and the unlabelled data from the validation and test set. Two different classifiers are used, each for both the domains. Discrepancy between the probability distributions between both the domains are minimized in the shared latent space.}
    \label{fig:one-hot encoding}
\end{figure*}
\section{Related Works}In this section,we briefly review the literature on SAR image classification and domain adaptation techniques.
\textbf{SAR image classification} Classical approaches like Wishart classifiers\cite{dabboor2013unsupervised}, random fields and traditional machine learning techniques like SVMs\cite{zhao2001support}\cite{lardeux2009support} and random forest\cite{hansch2018skipping} have been used for SAR image classification.But these approaches depend on manually handcrafted low-level features.
Deep learning models have been used for SAR classification as they can extract features in an end-to-end manner.
Hansch \textit{et al.}\cite{hansch2009classification} used complex-valued multi-layered perceptron to classify complex-valued PolSAR images.
Yang \textit{et al.}\cite{Yang_2021_CVPR} used a two branch framework with  cascading and paralleling experts to classify SAR images with a long-tailed distribution.
Zhang \textit{et al.}\cite{8518829} proposed a deep CNN architecture called CompressUnit(CU) to classify minimal annotated high resolution SAR images.
A two stage training strategy was employed by Yibing \textit{et al.}\cite{9199258} wherein the first stage was used to train a CNN for normal classification task and in the second stage, the output of the middle layer of the network in the forward propagation process was extracted to train an end-to-end metric network to learn the relations between the sample features.\\
\textbf{Domain adaptation.}
Models trained for a particular domain-specific task experience a reduction in performance when testing on a different but related domain.
Domain shift can be broadly categorized into three major categories:(1)covariant shift,(2)concept shift and (3)prior probability shift\cite{lee2019sliced}.For concept drift,the discrepancies between feature distributions in target and source domain are minimized using metrics like Maximum Mean Discrepancy(MMD)\cite{gretton2009covariate} and KL divergence\cite{daume2006domain}.
Lee \textit{et al.}\cite{lee2019sliced} proposed an unsupervised learning algorithm using the Sliced-Wasserstein Distance(SWD) metric to measure the dissimilarity between two probability distributions.A geometrically meaningful guidance was provided for the target samples far from the source distribution for efficient alignment in an end-to-end manner. 
Heitz \textit{et al.}\cite{heitz2021sliced} showed the Sliced-Wasserstein Distance(SWD) is a superior alternative to Gram-matrix loss for measuring the distance between two distributions in the feature space for neural texture synthesis in terms of optimization or for training generative neural networks.
Rostami \textit{et al.}\cite{9025419} used SWD as a metric to minimize the discrepancy between source and target distributions as it is a differentiable metric with non-vanishing gradients problem\cite{redko2017theoretical} to transfer knowledge from source to target domain.\\

\section{Methodology}
This paper proposes Multi-Modal Domain Fusion(MDF) network for two-way transfer of knowledge between the EO and SAR domains. In the following, the overall framework and the problem formulation of the network are explained. The weighted data sampling and the training of the network are also discussed to deal with the long-tailed distribution of the challenge dataset.\\
\textbf{Overall framework} 
The overall framework of our proposed method is shown in Fig.\ref{fig:one-hot encoding} where twin Resnet-50 networks are used.
The feature encoder networks with Resnet-50 backbones are trained using the labeled and unlabelled data points using the Sliced Wasserstein Discrepancy(SWD) loss. The latent space of the two networks $L_{EO}$ and $L_{SAR}$ are shared to create a domain invariant space\cite{9025419},\cite{gretton2009covariate},\cite{redko2017theoretical} and effectively improve the knowledge transfer across both the networks.
The individual classifier parameters are learnt using the combined focal loss\cite{lin2017focal} of both the networks in a supervised manner. Finally, the outputs of the individual classifiers of the two networks are passed to a fusion block to predict the classification outputs by taking the weighted average of the two networks, where the weights $w_1$ and $w_2$ are learnable parameters learnt using the least-squares approach.
\begin{equation}
  \begin{aligned}
    Y = w_1*y_{EO} + w_2*y_{SAR}
  \end{aligned}
\label{equation:fusion}
\end{equation}

\textbf{Problem formulation} Let \textit{X}$\subset$ $\Re^{d}$ denote the domain space of EO and SAR input data\cite{9025419}.
For multi-class classification problem with 10 classes in both the domains,i.i.d data samples are drawn from the joint probability distribution of $(x_i,y_i)$\texttildelow$q(x,y)$ with a marginal probability distribution of $p(x)$ over \textit{X}.Since EO and SAR are different domains,their marginal probability distributions are different in the latent space.\\
The main objective of Multi-Modal Domain Fusion (MDF) network is to train a deep learning model $f_\theta$ : \textit{X} $\rightarrow$ \textit{Y} $\subset$ $\Re^k$ with the learnable parameters $\theta$,where  \textit{Y} is the label space with final output predictions.
EO deep learning model $f_\theta$(.) consists of a feature encoder $\phi_v(.): \textit{X} \rightarrow \textit{Z}$ made up of convolutional layers to encode the input data to the latent space followed by a dense layer $h_w(.):\textit{Z} \rightarrow \textit{Y}$ to map the features from the latent space to the label space,where $\theta$=\textit{(v,w)} denote the learnable parameters.
SAR deep learning model $f_{\theta}^{'}$(.) consists of an encoder $\psi_u(.): \textit{X} \rightarrow \textit{Z}$ which maps the input data to the latent space followed by a dense layer $h_i(.):\textit{Z} \rightarrow \textit{Y}$ to map the features to the label space,where $\theta^{'}$=\textit{(u,i)} denote the learnable parameters,as shown in the Fig.\ref{fig:one-hot encoding}.\\
Let $D_E=(X_E,Y_E)$ be the N labeled images in the EO domain,where $X_E=[x^E_1, .. ,x^E_N]\in \Re^{dxM}$ be the input images and $Y_E=[y^E_1, .. ,y^E_N]\in \Re^{kx1}$ be the corresponding labels. The unlabeled images from the validation and test phase are denoted by $D'_E=(X'_E)$.
Similarly the labeled data from SAR domain can be represented as $D_S=(X_S,Y_S)$ where $X_S\in \Re^{dxM}$ be the input images and $Y_S\in \Re^{kx1}$ be the corresponding labels.The unlabelled images are from the test and validation phase are represented as $D'_S=(X'_S)$.\\
\textbf{Loss function:}The following loss function is proposed for computing the optimal values of $\theta$ and $\theta'$:

\begin{equation}
  \begin{aligned}
    \min_{\theta,\theta'} & \frac{1}{N}\sum_{i=1}^N {\textit{L}(h_w(\phi_v(x_i^{E});y_i^{E}))} \\
    & +\frac{1}{N}\sum_{i=1}^N {\textit{L}(h_i(\psi_u(x_i^{S});y_i^{S}))} \\
    & +\lambda{\textit{D}(\phi_v(p(X'_E)),\psi_u(p(X'_S)))} \\
    & +\eta \sum_{j=1}^k{\textit{D}(\phi_v(p(X_E)|y^E_j),\psi_u(p(X_S|y^S_j)))}\\
  \end{aligned}
\label{equation:loss}
\end{equation}
where \textit{L} is a loss function,\textit{D} is a discrepancy measurement metric and $\lambda,\eta$ are hyper-parameters. 
The first two terms of the loss function in equation(2) are used to train the feature encoders and the classifiers of EO and SAR models in a supervised learning fashion to learn the discriminating features of their respective domains.The loss function L used in equation(2) is the focal loss\cite{lin2017focal} as it is more suitable for imbalanced dataset compared to cross-entropy loss.\\
The discrepancy metric \textit{D} is the Sliced Wasserstein Distance(SWD) and is used to train the only the feature extractors.
The third term in equation(2) is the matching loss function.It is used to align the marginal probability distribution $p(X'_E)$ and $p(X'_S)$ of the unlabeled data samples  from the validation and test phase of both the domains in the latent space.The last term of the loss function is used to align the class-conditional probabilities of the labeled training data of both the domains in the latent embedding space to maintain the semantic consistency.
Labelled and unlabelled images of EO and SAR are used to learn discriminative domain invariant features.

\textbf{Weighted Data Sampling} The long-tailed nature of the PBVS-MAVOC challenge dataset makes it very hard for the network to recognize and predict the tail class images. The network over-fits on the head classes as the number of images in the head classes are approximately more than 10 times the number of images in the tail classes. Long-tailed representation problems are therefore addressed mainly through data re-balancing, data re-sampling\cite{4717268}, re-weighting\cite{cui2019classbalancedloss}, data augmentation\cite{Kim_2020_CVPR} and two-stage training methods\cite{Kang2020DecouplingRA}. Therefore in our work, several image augmentation techniques were used for the tail classes. 
A training dataset for both the domains has been manually curated with around 7000 randomly selected images per class for head classes and 5500 images per class for the tail classes using augmentation techniques like random rotation, horizontal and vertical flips.
To account for the slight class imbalance in the manually curated dataset, a weighted random sampler was used with the weights(probability of an image being picked from a class) being equal to 1/$n_i$, where $n_i$ refers to the number of samples in the $i^{th}$ class.\\

\section{Experiments}
In this section, the challenge dataset and the implementation details of the MDF network are discussed followed by the quantitative evaluation of the network on the dataset.An ablation study is conducted to show the effectiveness of the MDF network.\\
\textbf{Dataset:}
The PBVS 2022 MAVOC challenge is on the NTIRE 2021 Multi-Aerial View Object Classification dataset\cite{Liu_2021_CVPR} which is a long-tailed dataset with the head classes dominating the percentage(\%) of training samples when compared to the tail classes as shown in Table \ref{table:dataset}. 
The dataset consists of images taken from several Electro-optical(EO) and SAR sensors mounted on an airplane. The SAR images are of higher resolution than the EO images. The SAR images are of size 55X55 pixels compared to the EO images that are of size 31X31 pixels. Table \ref{table:dataset} displays the number of training samples in each class. Few training data samples from the EO domain and the SAR domain are shown in Fig.\ref{fig:dataset}.
\begin{table}[htb!]
\begin{adjustbox}{width=0.98\linewidth}
\begin{tabular}{|l|l|l|l|}
\hline
\textbf{Class \#} & \textbf{\begin{tabular}[c]{@{}l@{}}Class\\ Name\end{tabular}} & \textbf{\# Train Samples} & \textbf{\%  of samples} \\ \hline
0 & sedan & 234,209 & 79.72\% \\ \hline
1 & suv & 28,089 & 9.56\% \\ \hline
2 & pickup truck & 15,301 & 5.20\% \\ \hline
3 & van & 10,655 & 3.62\% \\ \hline
4 & box truck & 1,741 & 0.59\% \\ \hline
5 & motorcycle & 852 & 0.29\% \\ \hline
6 & flatbed truck & 828 & 0.281\% \\ \hline
7 & bus & 624 & 0.212\% \\ \hline
8 & \begin{tabular}[c]{@{}l@{}}pickup truck\\ with trailer\end{tabular} & 840 & 0.286\% \\ \hline
9 & \begin{tabular}[c]{@{}l@{}}flatbed truck\\ with trailer\end{tabular} & 633 & 0.215\% \\ \hline
 &  & Total = 293,772 &  \\ \hline
\end{tabular}
\end{adjustbox}
\caption{ PBVS 2022 MAVOC challenge Long-Tailed Training Samples Distribution.}
\label{table:dataset}
\end{table}
The challenge had two tracks,track-1 with only SAR test data and track-2 with EO and SAR test data.
The validation and test set have uniformly distributed samples among all the classes with the validation set having 770 images in total per domain and test set having 826 images in total per domain.

\subsubsection{Implementation}
A joint training strategy is adopted to train the twin Resnet-50\cite{he2016deep} networks, one for EO images and other for SAR images, using the given labeled data as well as the unlabeled data as shown in the Fig.\ref{fig:one-hot encoding}.
The input images are resized to 224x224 pixels before passing it to the Resnet-50 models.
Data is loaded to the  respective networks such that the resized EO image and the resized SAR image are the images of the same object but in the two different domains.\\
A two-stage training approach has been adopted where a Resnet-50\cite{he2016deep}  network is trained on the original imbalanced EO images of the challenge dataset for 15 epochs using focal loss\cite{lin2017focal} as it gives more importance to the hard samples and takes care of the class imbalance to a certain extent. The pretrained weights are used in the SAR Resnet-50  network of the MDF architecture for a better initialization. This initialization was not used for the EO encoder as it tended to over-fit on the dataset.
MDF network is trained on the manually curated dataset with the weighted random sampler for 100 epochs with a batch size of 64.
A learning rate scheduler is used to make the networks learn better with the initial setting being equal to 0.03 using Adam Optimizer.\\ 
In track-1,the Resnet-50 backbones are replaced with the EfficientNetb0\cite{tan2019efficientnet} in the MDF network and are trained for  50 epochs with batch size of 32.
The weighted average of the Resnet-50 based SAR model and EfficientNetb0 based SAR model as per Equation \ref{equation:fusion} are taken for making final predictions.\\
In track-2, only the Resnet-50 backbone based MDF network is used to make final predictions as the inclusion of EfficientNetb0 in track-2 did not improve the results.
The implementation was done using Pytorch on Nvidia 3090Ti GPU.
\subsubsection{Experimental results}
Table \ref{table:track2} shows the test results on the PBVS 2022 MAVOC challenge - Track 2,where both the sensory information has been provided. Our MDF network places us in the top-5 showing that using both the sensory information helps better the accuracy of aerial view object classification. \\
\begin{table}[htb!]
\centering
\begin{tabular}{|l|l|l|}
\hline
\textbf{\# Place} & \textbf{Team} & \textbf{\begin{tabular}[c]{@{}l@{}}Top-1\%\\ Accuracy\end{tabular}} \\ \hline
1 & Team A & 51.09\% \\ \hline
2 & Team B & 46.85\% \\ \hline
3 & Team C & 41.77\% \\ \hline
4 & Team D & 37.65\% \\ \hline
\textbf{5} & \textbf{MDF} & \textbf{34.26\%} \\ \hline
\end{tabular}
\caption{ Test results for the PBVS 2022 MAVOC challenge in Track-2(SAR+EO) where the proposed MDF network is placed in the 5th place.}
\label{table:track2}
\end{table}
Table \ref{table:track1} shows the results on the track-1 SAR only test data. The proposed MDF approach places us in the top-10 with our result being comparable to the results of the top-5 placed teams demonstrating that the multi-source domain fusion approach can be used to effectively classify SAR data.

\begin{table}[htb!]
\centering
\begin{tabular}{|l|l|l|}
\hline
\textbf{\# Place} & \textbf{Team} & \textbf{\begin{tabular}[c]{@{}l@{}}Top-1\%\\ Accuracy\end{tabular}} \\ \hline
1 & Team A & 36.44\% \\ \hline
2 & Team B & 31.23\% \\ \hline
3 & Team C & 28.09\% \\ \hline
4 & Team D & 27.97\% \\ \hline
5 & Team E & 27.48\% \\ \hline
\textbf{9} & \textbf{MDF} & \textbf{25.30\%} \\ \hline
\end{tabular}
\caption{ Test results for the PBVS 2022 MAVOC challenge in Track-1(SAR) where our proposed approach placed us in the top-10(9th place).}
\label{table:track1}
\end{table}
We get better results in track-2 when compared to track-1 since the SAR images have  features that are hard to interpret compared to the EO images.
EO images are easier to learn and therefore the MDF network benefits from correctly classifying images in the SAR+EO dataset when compared to only SAR dataset.
\begin{table*}[h]
\centering
\begin{tabular}{|l|l|l|l|l|l|l|l|l|l|}
\hline
\textbf{\begin{tabular}[c]{@{}l@{}}No.\\ of\\ models\end{tabular}} & \textbf{\begin{tabular}[c]{@{}l@{}}Feature \\ extractor\end{tabular}} & \textbf{\begin{tabular}[c]{@{}l@{}}Pre-\\ trained\end{tabular}} & \textbf{\begin{tabular}[c]{@{}l@{}}Re-\\ sampled\end{tabular}} & \textbf{\begin{tabular}[c]{@{}l@{}}Augmen-\\ tation\end{tabular}} & \textbf{\begin{tabular}[c]{@{}l@{}}Supervised\\ Multi-\\ Domain\\ Fusion\end{tabular}} & \textbf{\begin{tabular}[c]{@{}l@{}}Semi-\\ Supervised\\ Multi-\\ Domain\\ Fusion\end{tabular}} & \textbf{\begin{tabular}[c]{@{}l@{}}Validation\\  set\end{tabular}} & \textbf{\begin{tabular}[c]{@{}l@{}}Test\\  set\end{tabular}} & \textbf{\begin{tabular}[c]{@{}l@{}}Top-1\% \\ Accuracy\end{tabular}} \\ \hline
Single & Resnet-50 & \begin{tabular}[c]{@{}l@{}}Yes\\ (on Image-net)\end{tabular} & - & - & - & - & Yes & - & 13.896\% \\ \hline
Single & Resnet-50 & \begin{tabular}[c]{@{}l@{}}Yes\\ (on EO data)\end{tabular} & Yes & - & - & - & Yes & - & 15.45\% \\ \hline
Single & Resnet-50 & \begin{tabular}[c]{@{}l@{}}Yes\\ (on EO data)\end{tabular} & Yes & Yes & - & - & Yes & - & 16.49\% \\ \hline
Twin & Resnet-50 & \begin{tabular}[c]{@{}l@{}}Yes\\ (on EO data)\end{tabular} & Yes & - & Yes & - & Yes & - & 17.01\% \\ \hline
Twin & Resnet-50 & \begin{tabular}[c]{@{}l@{}}Yes\\ (on EO data)\end{tabular} & Yes & Yes & Yes & - & - & Yes & 21.91\% \\ \hline
Twin & Resnet-50 & \begin{tabular}[c]{@{}l@{}}Yes\\ (on EO data)\end{tabular} & Yes & Yes & - & Yes & - & Yes & 24.57\% \\ \hline
Twin & \begin{tabular}[c]{@{}l@{}}Resnet-50,\\ Efficient\\ -Netb0\end{tabular} & \begin{tabular}[c]{@{}l@{}}Yes\\ (on EO data)\end{tabular} & Yes & Yes & - & Yes & - & Yes & 25.3\% \\ \hline
\end{tabular}
\caption{Validation phase and test phase results showing the importance of the proposed approach compared to the other various different settings on the validation and test data of the PBVS Multi-Modal Aerial View Object Classification-Track 1 SAR dataset.}
\label{table:ablation}
\end{table*}

\subsubsection{Ablation study}
Table \ref{table:ablation} shows experimental results on the validation and test SAR data for track 1 with various settings like  pretraining, data augmentation and comparison of MDF network with the baseline Resnet-50 model. The table also displays the effectiveness of using unlabeled data points along with the given labeled data points. The effectiveness of the joint training strategy of the twin network is displayed as the accuracy of the using supervised MDF increased the validation accuracy to 17.01\% on the original challenge dataset(no data augmentation) when compared to the baseline Resnet-50 network with data-augmentation and re-sampling. Using semi-supervised MDF network,i.e,training using labeled and unlabeled samples gave better results compared to the supervised MDF network on the test dataset. The accuracy increased from 21.91\% to 24.57\% when the unlabeled data points were used along with the labeled data points thereby empirically proving that the proposed training strategy makes good use of all the available data to reduce the domain invariance and classify the multi-modal aerial view objects accurately.\\
The long-tailed nature of the dataset causes the network to over-fit on the head classes. To overcome the over-fitting issue,various image augmentation techniques like rotation, horizontal flips, vertical flips, random crops and center crops are applied to the tail classes.
The baseline Resnet-50 model trained using this image augmented SAR dataset and the weighted random sampling increased the validation accuracy from 13.896\% to 16.49\%.
We also tried a shared classifier approach instead of using two individual classifiers for each domain but that did not give good results on the validation data.The poor performance maybe due to the long tailed nature of the dataset.\\
On the SAR test set, the MDF network trained using Cross entropy loss gave a top-1\% accuracy of 24.09\% and using Focal loss\cite{lin2017focal}, gave a top-1\% accuracy of 25.06\%. The focal loss gives more importance to the training samples that are hard to predict, and it giving a better accuracy than cross-entropy loss proves that some class samples are indeed harder to predict than some other class samples in the dataset.  The similar trend was observed even with SAR+EO image inputs. 
\section{Conclusion}
In this paper, we present a  Multi-Modal Domain Fusion network, consisting of twin networks followed by a fusion block to accurately classify SAR and EO images.Semi-supervised learning is used to train the network using labeled and unlabeled data samples.
The proposed method yields competitive results in both the tracks making our solution one of the top-ranked solutions in the PBVS 2022 MAVOC Challenge and therefore empirically proves the effectiveness of the approach.

{\small
\bibliographystyle{ieee_fullname}
\bibliography{egbib}
}

\end{document}